%% file: main_arxiv.tex
\documentclass[10pt,twocolumn]{article}

\usepackage[margin=0.72in]{geometry}
\usepackage{cmap}
\usepackage[T1]{fontenc}
\usepackage[utf8]{inputenc}
\input{glyphtounicode}
\usepackage{times}
\usepackage{microtype}
\usepackage{graphicx}
\usepackage{booktabs}
\usepackage{tabularx}
\usepackage{amsmath,amssymb,amsfonts}
\usepackage{mathtools}
\usepackage{bm}
\usepackage[numbers,sort&compress]{natbib}
\usepackage{xcolor}
\usepackage{hyperref}
\usepackage{caption}
\usepackage{subcaption}
\usepackage{multirow}
\usepackage{array}
\usepackage{float}
\usepackage{dblfloatfix}
\usepackage{fancyhdr}
\usepackage{makecell}
\usepackage{adjustbox}
\usepackage{placeins}
\usepackage{enumitem}
\setlist{nosep,leftmargin=*}
\hypersetup{
  colorlinks=true,
  linkcolor=blue,
  citecolor=blue,
  urlcolor=blue,
  pdftitle={Mechanism Learning: Prototype-Anchored Mechanism Inference for Scientific Forecasting},
  pdfauthor={Qian Jiang and Liping Sun}
}
\makeatletter
\g@addto@macro\UrlBreaks{\do\/\do\-}
\makeatother

\title{Mechanism Learning: Prototype-Anchored Mechanism Inference for Scientific Forecasting}
\author{
Qian Jiang\\
School of Computing\\
The Australian National University\\
Canberra, ACT 2601, Australia\\
\texttt{u8075121@anu.edu.au}
\and
Liping Sun\\
iHuman Institute\\
ShanghaiTech University\\
Shanghai 201210, China\\
\texttt{sunlp@shanghaitech.edu.cn}
}
\date{}

\begin{document}
\twocolumn[
\begin{@twocolumnfalse}
\maketitle
\begin{abstract}
Scientific forecasting typically relies on direct state prediction, an approach that grows brittle under data scarcity, extended horizons, non-stationary dynamics, or high-dimensional complexity. While raw state trajectories are highly sensitive in these regimes, underlying local evolution rules often exhibit robust reusability. We introduce \emph{mechanism learning}, a framework that forecasts future states by estimating the currently active local mechanism. Our method compresses local spatiotemporal fragments into mechanism descriptors, forming a data-driven, structured mechanism space where proximity reflects similar local evolution rules. To ground these estimates in observed data, we utilize prototype anchors, a set of representative mechanisms that sparsely cover the space of local rules. We evaluate this approach on Burgers dynamics, WeatherBench2, and Lorenz96. Empirically, the learned mechanism spaces resist collapse and maintain strong local consistency. Compared to direct prediction and other models including FNO, NODE, LSTM, and reservoir-family methods, our framework demonstrates predictive gains in fragile regimes: it significantly improves switching stability in Burgers dynamics and achieves state-of-the-art performance both under the scarce-data fixed-horizon WeatherBench2 protocol and in intermediate-complexity Lorenz96. Ablation studies and drift diagnostics confirm that these improvements are driven by finite prototype anchoring rather than sheer latent capacity. Together, these results establish \emph{mechanism learning} as a principled, robust alternative to direct state prediction in forecasting complex systems.
\end{abstract}
\vspace{1em}
\end{@twocolumnfalse}
]

\section{Introduction}

Scientific forecasting can be organized by the predictive coordinate through which an observed history is mapped to the future. In direct state prediction, the map is \(H_t \mapsto x_{t+\Delta}\): recent states are mapped directly to future states. Operator-learning and latent-dynamics approaches instead map histories into learned operators, latent states, or lifted coordinates before prediction~\citep{Li2021FNO,Lu2021DeepONet,krishnan2015deep,korda2018linear}. In this paper we study a different intermediate coordinate: local spatiotemporal fragments are mapped to mechanism descriptors, these descriptors populate an empirical mechanism space, and the current history is forecast through the inferred active local mechanism.

Existing structured forecasting methods address these difficulties from several directions. Some models impose or exploit known dynamical structure, such as continuous-time dynamics or equation-aware weather models~\citep{Chen2018NeuralODE,Kochkov2024NeuralGCM}. Others learn operators, latent state-space models, or lifted dynamical coordinates~\citep{Li2021FNO,Lu2021DeepONet,krishnan2015deep,korda2018linear}. A complementary line extracts data-driven dynamical structure through modal decompositions or sparse equation discovery~\citep{schmid2010dynamic,brunton2016discovering}. Prototype and memory-based methods provide reusable supports for prediction or classification~\citep{Vinyals2016MatchingNetworks,Snell2017ProtoNets}. These directions introduce structure beyond raw state regression, but they typically do not treat the \emph{currently active} local mechanism as the online object of inference, nor as an explicit pipeline from observed history to inferred mechanism to future state.

We introduce \emph{mechanism-space forecasting}: instead of learning only a map from observed histories to future states, the model first infers a local mechanism descriptor \(\hat{\theta}_t\), locates it in an empirical mechanism space, and forecasts through that inferred coordinate. Here, \emph{local} refers to how descriptors are estimated from short spatiotemporal fragments or neighborhoods, while \emph{short-term} refers to the finite forecast window over which the descriptor is used. The resulting mechanism space is empirical: it is a predictive geometry constructed and diagnosed from data, not an assertion that the learned coordinates are a complete chart of true physical degrees of freedom.

Our contribution is to make this missing object explicit. We study whether empirical mechanism spaces are non-degenerate, locally coherent, and sparsely coverable; whether mechanism-space inference improves stability under switching dynamics and longer leads; and whether the advantage depends on anchored mechanism support rather than an arbitrary latent bottleneck. The central question is when the local mechanism provides a more stable predictive coordinate than the observed state trajectory.

\paragraph{Contributions.}
We make four contributions. \textbf{Mechanism-space forecasting:} we formulate scientific forecasting as online inference of the currently active local mechanism, rather than only direct state prediction. \textbf{Operational diagnostics:} we provide testable diagnostics for non-collapse, local coherence, and sparse coverability across Burgers dynamics, WeatherBench2, and Lorenz96. \textbf{Predictive gains in fragile regimes:} we show that the Full model improves switching stability in Burgers dynamics, achieves state-of-the-art performance under the scarce-data fixed-horizon WeatherBench2 protocol, and supports the intermediate-complexity picture in Lorenz96. \textbf{Evidence for prototype anchoring:} NoBank ablations and drift diagnostics show that the gains are tied to finite prototype anchoring rather than merely inserting a generic latent pathway.

Our empirical study spans three systems with complementary roles (Table~1). WeatherBench2 is the primary real-world benchmark and the primary external (FNO) testbed. Burgers provides sparse coverability, switching, and internal mechanism-space evidence. Lorenz96 provides a controlled chaotic system for sweet-spot sweeps, with LSTM and NODE baselines.

A fuller comparison to direct forecasting, latent dynamics and operator learning, prototype/retrieval methods, and benchmark settings \citep{Rasp2024WeatherBench2,Takamoto2022PDEBench,Lorenz1996Predictability} is provided in Appendix~\ref{app:extended-related-work} (Related Work).

% =====================================================
% ALL MAIN TEXT FIGURES MUST APPEAR BEFORE REFERENCES
% =====================================================

\begin{figure*}[t]
\centering
\includegraphics[width=0.95\textwidth]{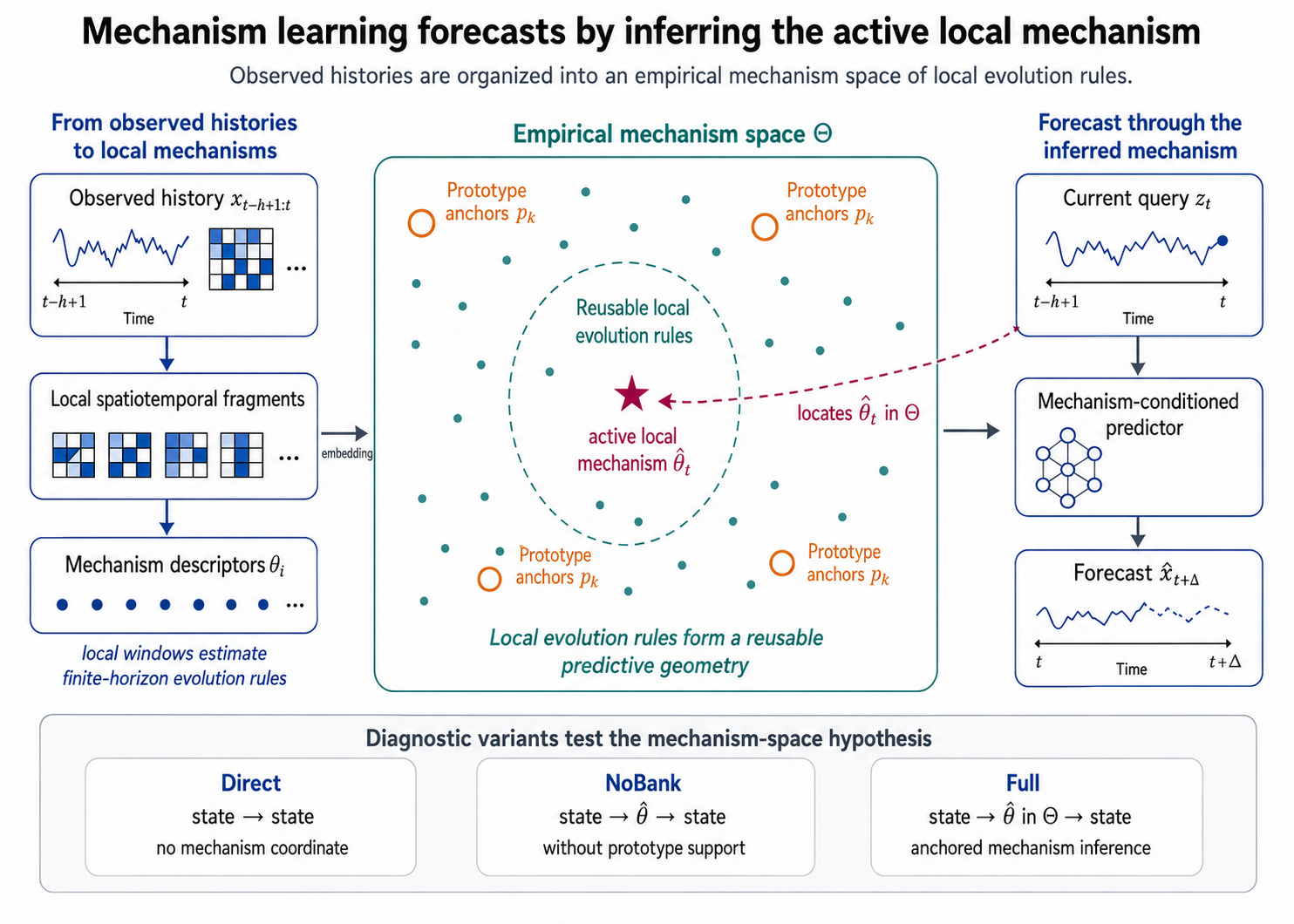}
\caption{%
\textbf{Mechanism learning forecasts by inferring the active local mechanism.}
Observed histories are decomposed into local spatiotemporal fragments and converted into mechanism descriptors \(\theta_i\), which populate an empirical mechanism space \(\Theta\) of local evolution rules.
At inference time, the current history produces a query \(z_t\) that locates the active mechanism \(\hat{\theta}_t\) within this space; forecasting then proceeds through a mechanism-conditioned predictor.
Prototype anchors \(p_k\) provide finite support inside the empirical geometry rather than serving as the main conceptual object.
The Direct, NoBank, and Full variants test whether predictive gains arise from forecasting through an anchored mechanism coordinate rather than from a generic intermediate latent pathway.%
}
\label{fig:1}
\end{figure*}

\section{Method}

\subsection{Problem Setup}

We consider forecasting in dynamical systems from partial or full observations of a time-evolving state. Let $x_t \in \mathbb{R}^{d_s}$ denote the observed system state at time $t$, and let the forecasting task be to predict a future state $x_{t+\Delta}$ from recent history. The standard forecasting paradigm is \emph{direct state prediction}, which learns a mapping
\[
f_{\mathrm{dir}}:\; (x_{t-h+1},\dots,x_t)\mapsto x_{t+\Delta}.
\]
This approach treats the forecasting problem as direct regression in state space.

In contrast, mechanism-space forecasting introduces an intermediate predictive coordinate, $\theta_t \in \mathbb{R}^{d_m}$, intended to summarize the local evolution rule active around time $t$. Forecasting then proceeds in two stages:
\[
(x_{t-h+1}, \ldots, x_t) \mapsto \hat{\theta}_t \mapsto \hat{x}_{t+\Delta}.
\]
The first arrow is mechanism inference: it locates the current dynamics within an empirical mechanism space. The second arrow is mechanism-conditioned prediction. The distinction is therefore not simply architectural. In the direct paradigm, forecasting is performed in state space; in the mechanism-space paradigm, forecasting is routed through a structured geometry of local evolution rules.

\paragraph{Operational mechanism-space objects.}
Let \(H_t=(x_{t-h+1},\ldots,x_t)\) denote the recent observed history, and let \(W_i\) denote a local spatiotemporal fragment or neighborhood extracted from the training system. A system-specific extraction map \(\psi\) produces a local mechanism descriptor
\[
\theta_i=\psi(W_i)\in\mathbb{R}^{d_m},
\]
whose role is to summarize the local evolution rule used for prediction over the finite lead \(\Delta\). The empirical mechanism space is the sampled geometry
\[
\Theta=\{\theta_i\}_{i=1}^{N}\subset\mathbb{R}^{d_m}.
\]
We use \(\Theta\) in a task-level predictive sense: nearby points are intended to represent reusable local dynamics, not necessarily distinct true physical degrees of freedom. At inference time, the current history is encoded into a query \(z_t=E(H_t)\). Finite support points \(P=\{p_k\}_{k=1}^{K}\subset\mathbb{R}^{d_m}\) anchor inference within this empirical geometry, producing
\[
\alpha_t=\mathrm{Attn}(z_t,P),\qquad
\hat{\theta}_t=\sum_{k=1}^{K}\alpha_{t,k}p_k,\qquad
\sum_{k=1}^{K}\alpha_{t,k}=1.
\]
The forecast is then generated by a mechanism-conditioned predictor
\[
\hat{x}_{t+\Delta}=G(z_t,\hat{\theta}_t).
\]

Our motivating hypothesis is that there are scientific forecasting settings in which the mechanism representation is a more stable and reusable predictive coordinate than the raw state trajectory itself. Figure~1 summarizes the current implementation.

\subsection{Local Mechanism Extraction}

The first step in our framework is to instantiate the descriptor map \(\psi\) for each system. Given a local spatiotemporal fragment of data, the extraction procedure returns a descriptor \(\theta_i\) that summarizes the local dynamical behavior associated with that fragment. Intuitively, $\theta_i$ is not intended to encode the full state. It describes the local rule associated with a fragment: \emph{local} refers to the fragment, window, or neighborhood from which the descriptor is obtained, while \emph{short-term} refers to the finite forecast interval over which that descriptor is used to organize prediction.

The exact extraction procedure is system-dependent, because the observable structure and local fitting assumptions differ across Burgers, WeatherBench2, and Lorenz96. In all cases, however, the same conceptual role is preserved: each $\theta_i$ is obtained through a system-specific local mechanism extraction or parameterization step and is treated as a candidate point in mechanism space. Collecting many such representations from training data yields a set
\[
\Theta = \{\theta_1,\theta_2,\dots,\theta_N\},
\]
which serves as an empirical sample of the mechanism space supported by the system.

This grounding in local dynamical structure distinguishes mechanism learning from a generic latent bottleneck and makes the intermediate representation a meaningful object of geometric and physical analysis.

\subsection{Empirical Mechanism Space and Prototype Support}

The sampled set $\Theta$ defined above is used as an empirical mechanism space: a geometry of local evolution rules supported by the training system. Rather than treating this space as an unconstrained continuum at test time, we represent its reusable support through a finite prototype bank
\[
P = \{p_1,\ldots,p_K\}, \quad p_k \in \mathbb{R}^{d_m},
\]
where each prototype represents a reusable region of the empirical mechanism space.

Prototype support serves two purposes. First, it provides a sparse cover of the mechanism space. If the mechanism space is genuinely structured, then nearby local mechanisms should be reusable across samples, and forecasting should not require memorizing every extracted $\theta_i$ independently. Second, the finite support acts as a geometric prior: instead of allowing mechanism inference to wander arbitrarily in $\mathbb{R}^{d_m}$, prediction is anchored to previously observed and organized mechanism templates.

In practice, the prototypes may be selected, learned, clustered, or otherwise derived from the collection $\Theta$, depending on the implementation used in each system. What matters conceptually is that the mechanism space is not treated as an unconstrained latent continuum at test time. Instead, it is regularized through a finite set of representative mechanism patterns. This makes the framework explicitly different from standard latent-variable forecasting, in which the intermediate representation is typically free to drift without reference to a structured support set. Conceptually, the bank stores support points in an empirical mechanism geometry rather than a generic feature memory; the exact parameterization is implementation-dependent and summarized in Appendix A.

\subsection{Anchored Mechanism Inference}

At test time, the model must infer the current mechanism from observed state history. Let
\[
z_t = E(x_{t-h+1},\dots,x_t)
\]
denote a learned state-dependent query representation produced by an encoder $E$. The goal is not to map $z_t$ directly to the future state, but to infer an anchored mechanism estimate $\hat{\theta}_t$ by retrieving from the prototype bank.

Concretely, the query representation interacts with the prototype bank to produce weights
\[
\alpha_t = (\alpha_{t,1},\dots,\alpha_{t,K}), \qquad \sum_{k=1}^K \alpha_{t,k}=1,
\]
which indicate how strongly the current dynamics align with each learned prototype. The inferred mechanism is then obtained by a structured aggregation
\[
\hat{\theta}_t = \sum_{k=1}^K \alpha_{t,k} p_k,
\]
or by an equivalent anchored retrieval rule used by the particular implementation. The key idea is that $\hat{\theta}_t$ is not predicted freely from scratch; it is inferred through structured retrieval over learned mechanism prototypes.

The future state is then predicted through a mechanism-conditioned predictor
\[
\hat{x}_{t+\Delta} = G(z_t,\hat{\theta}_t),
\]
where $G$ uses both the encoded state context and the inferred mechanism; system-specific details are given in Appendix A. This decomposition is the defining property of the framework: forecasting is carried out through a mechanism estimate that is explicitly constrained by the geometry of the learned mechanism space.

\section{Results}

We organize the results as an evidence chain for mechanism-space forecasting. First, we test whether empirical mechanism spaces behave as structured objects rather than arbitrary hidden layers. Second, we ask whether forecasting through such spaces improves stability when local dynamics change. Third, we test whether the clearest real-system gains occur under scarce-data and longer-lead WeatherBench2 conditions. Fourth, NoBank and drift diagnostics test whether the gain comes from anchored mechanism inference rather than a generic latent pathway. Controlled Lorenz96 experiments provide a complementary chaotic-system test, while full-data horizon scans and phase-sweep boundary analyses are reported in the appendix.

\subsection{Direct Baseline and Fair Comparison}

To evaluate whether mechanism learning provides a genuine forecasting advantage, we compare it against a matched \emph{direct state prediction} baseline. The baseline receives the same input history and predicts the future state directly, without routing through a mechanism space:
\[
\hat{x}_{t+\Delta}^{\mathrm{dir}} = G_{\mathrm{dir}}(E_{\mathrm{dir}}(x_{t-h+1},\dots,x_t)).
\]

Our comparisons are designed to be as fair as possible. Whenever applicable, we align the direct and mechanism-learning models along the same input/output task, data protocol, training protocol, paired seeds, and evaluation metrics. This design matters because any observed gain should be attributable to the structure of mechanism learning rather than to unequal training conditions.

\subsection{NoBank Ablation}
\label{sec:nobank}

A central question is whether the gain of mechanism learning comes from the existence of an intermediate mechanism pathway in general, or specifically from \emph{anchored inference over finite prototype support}. To isolate this issue, we introduce a \emph{NoBank} ablation.

The NoBank model retains the same overall mechanism-learning pipeline up to the intermediate mechanism representation, but removes prototype retrieval over finite prototype support. Instead of inferring $\hat{\theta}_t$ through structured retrieval over prototypes, the model predicts $\hat{\theta}_t$ directly from the encoded state representation:
\[
\hat{\theta}_t^{\mathrm{NoBank}} = H(z_t),
\]
where $H$ is a learned mapping from the query representation to the mechanism representation. The future state is then predicted through the same mechanism-conditioned predictor:
\[
\hat{x}_{t+\Delta}^{\mathrm{NoBank}} = G(z_t,\hat{\theta}_t^{\mathrm{NoBank}}).
\]

This ablation is deliberately precise. It does not remove the mechanism pathway altogether. It preserves the intermediate mechanism representation and mechanism-conditioned prediction, while deleting only finite prototype support and its anchoring effect. As a result, it directly tests whether the gains of the Full model arise from routing through any mechanism-like latent variable, or from routing through a mechanism estimate that is constrained by a structured prototype space.

\begin{table*}[t!]
\centering
\small
\setlength{\tabcolsep}{4pt}
\renewcommand{\arraystretch}{1.12}
\caption{Mechanism construction and the role of each benchmark. \emph{WeatherBench2 uses fixed-horizon supervised evaluation, not autoregressive rollout.} Lorenz96 rollouts in Appendix~\ref{app:external} are autoregressive \emph{evaluation} under the one-step training setup.}
\label{tab:mechanism_construction}
\begin{tabularx}{\textwidth}{@{}l >{\raggedright\arraybackslash}X >{\raggedright\arraybackslash}X @{}}
\toprule
\textbf{System} & \textbf{Mechanism construction} & \textbf{Why it matters here} \\
\midrule
Burgers &
Local fit on spatiotemporal windows (regularized predictor); prototypes \emph{via} selection or clustering. &
Local PDE grounding; sparse cover, switching, and internal mechanism-space evidence. \\
\midrule
WB2 &
Encoder-driven retrieval from a bank of recent $T$ and $Z500$ history. &
Primary real benchmark and external-FNO line under the fixed-horizon supervised protocol. \\
\midrule
L96 &
Learned bank from $h{=}4$ windows; one-step training; AR test at $h\!\in\!\{1,2,4,8,16\}$; vs.\ LSTM and continuous-time NODE. &
Sweet-spot scans over $(F,N)$; discrete vs.\ continuous-time baselines. \\
\bottomrule
\end{tabularx}
\end{table*}

\subsection{Experimental Design, Benchmarks, and Baselines}
\label{sec:experimental-design}

We evaluate on three systems with fixed roles (Table~1). WeatherBench2 is the primary real-world benchmark and the primary external-baseline (FNO) testbed for the main WB2 tables. Burgers targets sparse cover and switching stability and provides internal mechanism-space / NoBank-style evidence. Lorenz96 is the controlled chaotic system for sweet-spot sweeps, with both a discrete (LSTM) and a continuous-time (NODE) baseline.

Across experiments, we compare the Full model to a matched Direct predictor and, where relevant, NoBank and an external FNO trained under the same fixed-horizon supervised WeatherBench2 protocol for W1/W2, rather than under a free-running autoregressive rollout protocol. All key comparisons follow matched task, data, metric, and paired-seed design \emph{when applicable}; the external FNO line is an additional baseline, not a replacement for the matched internal Direct-vs-Full comparison.
We report trainable parameter counts separately for transparency; our comparisons are matched in task, data protocol, metrics, and evaluation setup, but are not claimed to be strictly parameter-matched in every system.

\subsection{Empirical mechanism spaces are structured and sparsely sampleable}

The first question is whether the learned intermediate coordinate behaves as a meaningful empirical mechanism space, rather than an arbitrary latent artifact of the forecasting pipeline. Across all three systems, the extracted mechanism representations are non-degenerate, exhibit neighborhood continuity, and organize into structured geometries rather than collapsing into uninformative clouds. In Burgers, they can also be effectively covered by a finite number of prototypes, indicating that the mechanism space is not only structured but also sparsely sampleable. Taken together, these diagnostics provide operational evidence that mechanism-space forecasting is not built on an arbitrary hidden layer, but on an empirically recoverable predictive geometry.

These diagnostics play a specific role in the argument: they are operational tests for whether the learned intermediate coordinate behaves as an empirical mechanism space, not merely auxiliary representation statistics. Burgers tests sparse coverability, asking whether a finite prototype support can cover the useful regions of the learned mechanism geometry. WeatherBench2 tests local neighborhood coherence, asking whether temporally or dynamically nearby states occupy nearby regions of mechanism space more often than random pairs. Lorenz96 tests non-collapse and local continuity, asking whether the learned mechanism coordinates vary nontrivially while preserving neighborhood structure. Taken together, these diagnostics operationalize the claim that the intermediate coordinate is not an arbitrary latent layer, but an empirical predictive geometry of local evolution rules. Appendix~\ref{app:burgers-theta-umap} provides qualitative 2D UMAP summaries of the $K{=}512$ Burgers prototype $\theta$-bank; these panels are illustrative complements to the quantitative mechanism-space diagnostics in the main text.

\begin{figure*}[t]
\centering
\includegraphics[width=0.95\textwidth]{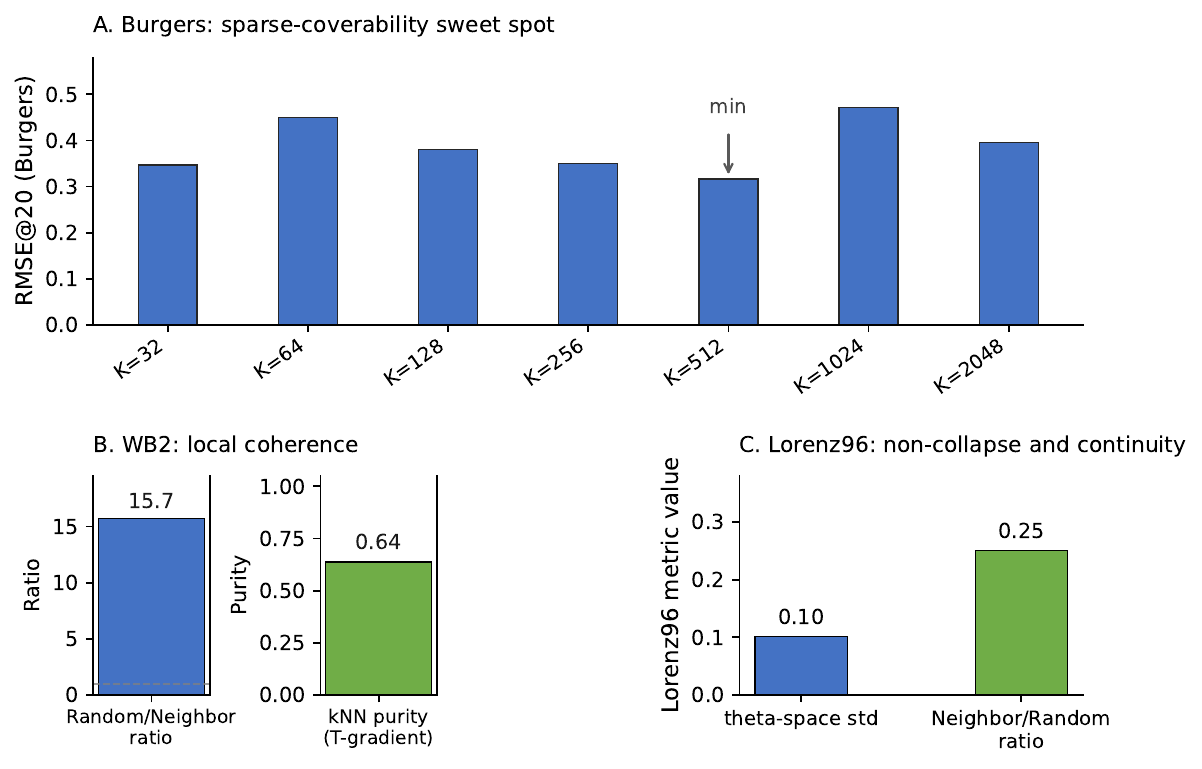}
\caption{Operational mechanism-space diagnostics. Burgers tests sparse coverability through prototype support size; WeatherBench2 tests local neighborhood coherence; Lorenz96 tests non-collapse and local continuity. Together, these diagnostics evaluate whether the intermediate coordinate behaves as an empirical predictive geometry rather than an arbitrary latent layer.}
\label{fig:2}
\end{figure*}

Burgers provides the clearest evidence for sparse coverability. Varying the prototype support size reveals a clear sweet spot over the displayed values $K\in\{32,64,128,256,512,1024,2048\}$: $K=512$ achieves the best displayed performance (near-split RMSE@20 $=0.32$), while larger support sizes such as $K=1024$ and $K=2048$ remain worse than the sweet-spot setting. On WeatherBench2, mechanism representations remain highly structured: temporally adjacent samples are far closer in mechanism space than random pairs (mean neighbor distance $=3.51\times 10^{-4}$ versus random distance $=5.52\times 10^{-3}$), and k-nearest-neighbor purity with respect to physically meaningful proxies is substantially above chance (e.g., $0.64$ for temperature-gradient quantiles and $0.62$ for $z500$-gradient quantiles). Lorenz96 shows the same qualitative pattern: the learned mechanism coordinates are non-degenerate ($\theta$-space standard deviation $=0.10$) and exhibit strong continuity (neighbor/random ratio $=0.25$). Figure~\ref{fig:2} summarizes these results.

\subsection{Mechanism inference is more stable under switching}

\begin{figure*}[t]
\centering
\includegraphics[width=0.90\textwidth]{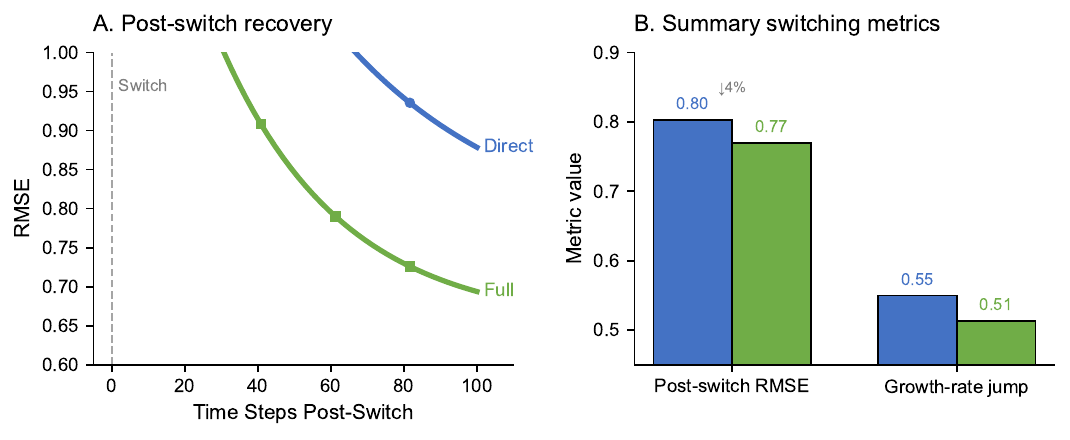}
\caption{Mechanism inference is more stable under switching. (A) Post-switch recovery after the regime change. (B) Summary switching metrics. The Full model exhibits lower post-switch RMSE and a smaller growth-rate jump than direct state prediction.}
\label{fig:3}
\end{figure*}

The first empirical advantage appears as improved stability under changing dynamics. In Burgers, this stability advantage is most visible under explicit switching. When the governing mechanism changes during prediction, mechanism inference exhibits systematically smaller post-switch error jumps, slower error growth, and smoother recovery than direct state prediction. Under the strong-switch case, post-switch RMSE decreases from $0.80$ for direct prediction to $0.77$ for mechanism inference, and the growth-rate jump at the switch decreases from $0.55$ to $0.51$ (Figure~\ref{fig:3}).

These results indicate that once the local rule is no longer stationary, reasoning through a structured mechanism space provides a more stable basis for extrapolation than relying on a single state-to-state map.

\subsection{Under scarce data and longer horizons, mechanism inference outperforms direct state prediction}

Under the current fixed-horizon scarce-data WeatherBench2 setting, the clearest real-system evidence appears at $+72$h: the Full model is ahead of the matched Direct and NoBank baselines, and it is also ahead of the external FNO baseline in the paired summary. This is the main real-system result of the paper. The easier $+24$h setting is reported separately in the appendix as a scope analysis (Appendix~\ref{app:wb2-external-baseline}).

\begin{figure*}[t]
\centering
\includegraphics[width=0.90\textwidth]{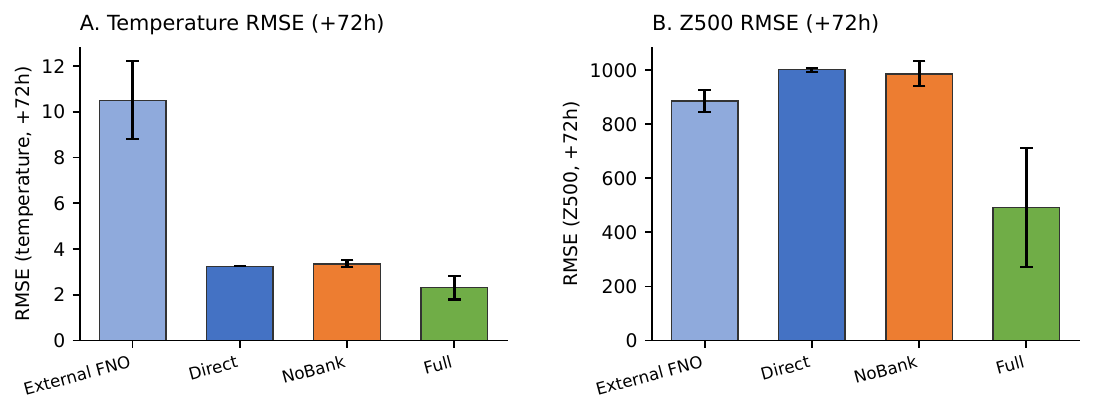}
\caption{\textbf{WeatherBench2 scarce-data \(+72\)h performance under fixed-horizon supervised evaluation.} The figure reports mean \(\pm\) standard deviation over five seeds for the four methods in Table~\ref{tab:wb2_summary}: External FNO, Direct, NoBank, and Full. The Full mechanism-space model gives the clearest real-system advantage at the harder \(+72\)h lead on both near-surface temperature and Z500. Seed-level paired tests and the easier \(+24\)h scope analysis are reported in Appendix~\ref{app:wb2-external-baseline}.}
\label{fig:wb2_72h_main}
\end{figure*}

Figure~\ref{fig:wb2_72h_main} summarizes mean $\pm$ standard deviation over seeds at $+72$h for External FNO, Direct, NoBank, and Full; Tables~\ref{tab:wb2_summary}--\ref{tab:wb2_paired} and seed-level statistics are in Appendix~\ref{app:wb2-external-baseline}. Because the Full model has substantially fewer trainable parameters than both the Direct and External FNO baselines in this WB2 setting (Table~\ref{tab:b3_wxpt}), the +72h advantage is not explained by a larger parameter budget; it points instead to the predictive coordinate introduced by mechanism-space inference under the controlled fixed-horizon protocol. Paired against External FNO at $+72$h, Full wins on all five seeds for temperature (two-sided $p \approx 5.5\times10^{-4}$) and on four of five seeds for Z500 ($p \approx 0.025$; Table~\ref{tab:wb2_paired}). The full-data horizon boundary scan is reported in Appendix~\ref{app:wb2-full-data-horizon-boundary}.

Appendix~\ref{app:wb2-style-baselines} reports additional same-protocol architecture-style stress baselines, including AFNO/FourCastNet-style, U-Net/ResNet-style, and GraphCast-style variants; these are low-data models trained and evaluated under the same fixed-horizon WeatherBench2 pipeline summarized in Table~\ref{tab:b1_wx}, not official pretrained GraphCast or Pangu systems and not global operational NWP benchmarks.

\subsection{Anchored mechanism inference is not a generic latent pathway}

NoBank retains the mechanism-conditioned predictor and the same overall pipeline, but removes finite prototype support so that $\hat{\theta}_t$ is predicted directly from the encoded state (Section~\ref{sec:nobank}). If NoBank matched Full, mechanism-space forecasting would reduce to a generic latent-bottleneck explanation.

\begin{figure*}[t]
\centering
\includegraphics[width=0.82\textwidth]{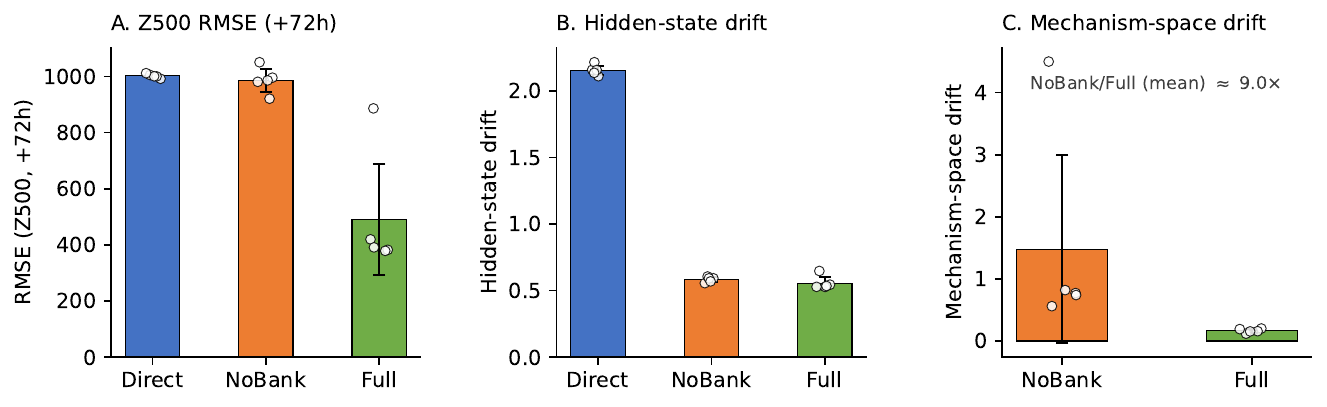}
\caption{NoBank and drift diagnostics distinguish anchored mechanism inference from a generic intermediate pathway. NoBank keeps the mechanism-conditioned predictor while removing empirical prototype support; drift summarizes hidden-state and mechanism-space stability. These diagnostics support the interpretation that the gain comes from anchored mechanism inference rather than from inserting an arbitrary intermediate pathway into the forecasting pipeline.}
\label{fig:6}
\end{figure*}

NoBank preserves the mechanism-conditioned pathway but removes finite prototype support. Its lack of improvement over Direct, together with the substantially smaller mechanism-space drift of the Full model (theta drift roughly $0.15$--$0.20$ versus $0.56$--$4.50$ for NoBank), supports the interpretation that anchored mechanism inference---not merely the presence of an intermediate latent pathway---drives the forecasting gain. Quantitatively, Full beats Direct on both temperature and $z500$ at +72h ($p=0.015$ and $p=0.006$, respectively), whereas NoBank shows no significant improvement over Direct ($p=0.18$ for temperature and $p=0.44$ for $z500$), and Full consistently outperforms NoBank (temperature $p=0.016$, $z500$ $p=0.006$). External-baseline comparisons, including the FNO line, are consistent with the same interpretation within the matched design (Appendix~\ref{app:external}). Figure~\ref{fig:6} summarizes the ablation and drift diagnostics.

\subsection{Controlled Lorenz96 supports the intermediate-complexity picture}

Controlled Lorenz96 experiments provide a complementary positive test in a chaotic system. In the tested phase sweeps, the largest gains occur at intermediate forcing and dimensionality, consistent with the view that mechanism-space forecasting is most useful when direct state extrapolation becomes brittle but local evolution rules remain reusable. In the sweet-spot setting \((F{=}8,N{=}16)\), the Full model remains strongest against Direct, LSTM, and NODE baselines in the reported autoregressive rollout comparison. Appendix~\ref{app:l96-reservoir-stress} adds an exact-protocol reservoir stress check against RC-ESN and NVAR, which further supports the state-of-the-art-level interpretation for this intermediate-complexity Lorenz96 setting. Detailed phase-sweep summaries and NODE comparisons are in Appendix~\ref{app:lorenz96-phase-sweep} and Appendix~\ref{app:node}.

% =====================================================
% DISCUSSION
% =====================================================

\section{Discussion}

\subsection{When mechanism space becomes the better predictive coordinate}

The results identify the conditions where mechanism-space forecasting becomes the stronger predictive coordinate: sparse data, longer leads, switching dynamics, and intermediate-complexity chaos. Appendix~\ref{app:external} and Appendix~\ref{app:wb2-full-data-horizon-boundary} record complementary scope refinements (including easier-lead WeatherBench2 behavior and Lorenz96 sweeps) that sharpen where the evidence is strongest without replacing the main chain.

Across experiments, the advantage appears as a coherent sequence: switching stability in Burgers, scarce-data $+72$h WeatherBench2 gains, and intermediate-complexity structure in controlled Lorenz96---each in settings where state trajectories become brittle while local evolution rules remain reusable.

These results suggest that, in difficult scientific forecasting, the limiting factor may be representational rather than merely architectural scale, which is precisely the regime where mechanism learning becomes relevant.

The evidence is concentrated in scientific forecasting settings where observed state trajectories become brittle: scarce data, longer leads, switching dynamics, and intermediate complexity. In these settings, mechanism-space forecasting becomes the better predictive coordinate.

\subsection{Why anchored support matters}

If only an intermediate vector mattered, NoBank would preserve the Full-model advantage. The ablation instead behaves as a falsification test for a generic latent pathway: removing finite support largely removes the gain, consistent with the drift panels in Figure~\ref{fig:6}. External baselines, including FNO on WeatherBench2, do not overturn this reading within the main matched design (Appendix~\ref{app:external}).

\section{Limitations}

\textbf{Mechanism-space interpretability.}
Mechanism space is treated here as an empirical predictive geometry, not as a guaranteed physically identifiable coordinate system. Our diagnostics show non-collapse, local coherence, and sparse coverability, but they do not prove that learned coordinates correspond uniquely to physical mechanisms. Future work should study when such geometries are stable across datasets, comparable across systems, or physically grounded, potentially drawing on complementary tools for modal analysis, lifted-coordinate dynamics, and sparse equation discovery~\citep{schmid2010dynamic,korda2018linear,brunton2016discovering}.

\textbf{Identifiability and formal definition.}
Our definition is operational: mechanisms are local descriptors useful for finite-horizon prediction under a given observation map and loss. Different local rules may induce indistinguishable finite-horizon forecasts, and the same predictive geometry may admit multiple coordinate representations. A formal theory should characterize predictive equivalence classes and conditions under which learned descriptors identify them.

\textbf{Scaling to extremely high-complexity systems.}
Our evidence is concentrated in controlled PDEs, a restricted WeatherBench2 fixed-horizon setting, and intermediate-complexity Lorenz96. Extremely high-complexity systems, full operational global forecasting settings, or open-ended nonstationary regimes may require richer mechanism libraries, stronger physical constraints, or additional scale-specific designs; reservoir approaches to large spatiotemporal chaos illustrate one relevant stress-test direction~\citep{pathak2018model}. Future work should test when mechanism-space inference remains useful as state complexity, variable count, and resolution increase substantially.

\textbf{Static prototype support.}
The present implementation uses finite prototype support learned from the training distribution. This support may be insufficient when new regimes appear outside the observed mechanism geometry. Adaptive, online, or regime-aware mechanism libraries are a natural next step.

% =====================================================
% CLEAR FLOATS BEFORE REFERENCES
% =====================================================

\clearpage
\FloatBarrier

% =====================================================
% REFERENCES
% =====================================================

\bibliography{references}

% =====================================================
% APPENDIX
% =====================================================

\clearpage
\appendix

\section{System-Specific Mechanism Extraction and Implementation Details}

\subsection{Burgers}

In the Burgers experiments, mechanism extraction is implemented as an explicit local fit over spatiotemporal windows. Let $\mathcal{W}(s,t)$ denote a local history window centered at location $s$ and time $t$, and let $\mathcal{N}_i$ denote the local neighborhood associated with anchor $i$. We define the mechanism descriptor $\theta_i$ by fitting a regularized local predictor:
\[
\theta_i
=
\arg\min_{\theta}
\sum_{(s,t)\in\mathcal{N}_i}
\mathcal{L}\!\bigl(P_{\theta}(\mathcal{W}(s,t)),\, y(s,t+\Delta)\bigr)
+
\lambda \|\theta\|_2^2.
\]

In our implementation, $P_{\theta}$ is a local linear / ridge predictor over the same input fragment used by the forecaster. The collection of extracted descriptors $\{\theta_i\}$ is then organized into a prototype bank
\[
P=\{p_1,\dots,p_K\},
\]
for example by clustering or representative selection. At test time, a query representation $z_t$ is computed from the current window, and anchored retrieval over prototype support yields
\[
\hat\theta_t=\sum_k \alpha_{t,k}p_k,
\]
which is then used by the downstream predictor.

\subsection{WeatherBench2}

For WeatherBench2, mechanism inference does not use an explicit local ridge-fit descriptor. Instead, the mechanism representation is produced through encoder-driven retrieval over a stored mechanism bank.

Let $\mathcal{W}_t$ denote the history window consisting of recent temperature and geopotential states, and let
\[
q_t = E_{\phi}(\mathcal{W}_t)
\]
be the query representation produced by the encoder. The mechanism bank is stored as a non-learnable buffer,
\[
\Theta^{\mathrm{bank}}=\{\theta^{\mathrm{bank}}_k\}_{k=1}^K,
\]
and retrieval produces the inferred mechanism
\[
\hat\theta_t=\sum_k \alpha_{t,k}\,\theta^{\mathrm{bank}}_k,
\]
where $\alpha_{t,k}$ is obtained by query-key attention over the bank entries.

Thus, WeatherBench2 is best viewed as a template-B variant: the mechanism is not obtained from an explicit local fit, but from encoder-driven retrieval over a stored bank of mechanism descriptors. The retrieved mechanism is then used by the predictor for $\Delta$-ahead forecasting. The choice between fixed and learnable banks is implementation-dependent and does not affect the core anchoring mechanism tested in this paper.

\subsection{Lorenz96}

For Lorenz96, the mechanism representation is implemented through a learned mechanism bank rather than an explicit local fit. Let $\mathcal{W}_t$ denote a local history window of length $h$, and let
\[
q_t = E_{\phi}(\mathcal{W}_t)
\]
be the query representation extracted from the current state history.

The mechanism bank is parameterized as a set of learnable entries,
\[
\Theta^{\mathrm{bank}}=\{\theta^{\mathrm{bank}}_k\}_{k=1}^K,
\]
with the bank implemented as trainable parameters. Prototype-anchored retrieval then gives
\[
\hat\theta_t=\sum_k \alpha_{t,k}\,\theta^{\mathrm{bank}}_k,
\]
where $\alpha_{t,k}$ is produced by attention over the bank entries.

In this sense, Lorenz96 instantiates template B: the mechanism is represented by a learned bank of reusable dynamical modes, and anchored retrieval constrains inference to remain close to previously organized mechanism patterns during chaotic rollout.

% =====================================================
% APPENDIX B (external / NODE / tabulated summaries)
% =====================================================
\input{appendix_arxiv}

\end{document}

%% file: appendix_arxiv.tex
% Included from mechanism_learning_neurips.tex (after Appendix A) via
%   \input{../paper_appendix/appendix}
% Contains: additional mechanism-space visual diagnostics; Appendix~B-style external baselines, protocol, NODE, and summary tables; extended related work.

\clearpage
\section{Additional Mechanism-Space Visual Diagnostics}
\label{app:mechanism-space-visual-diagnostics}

\subsection{Burgers theta-bank geometry}
\label{app:burgers-theta-umap}

\begin{figure}[!htbp]
    \centering
    \includegraphics[width=0.98\linewidth]{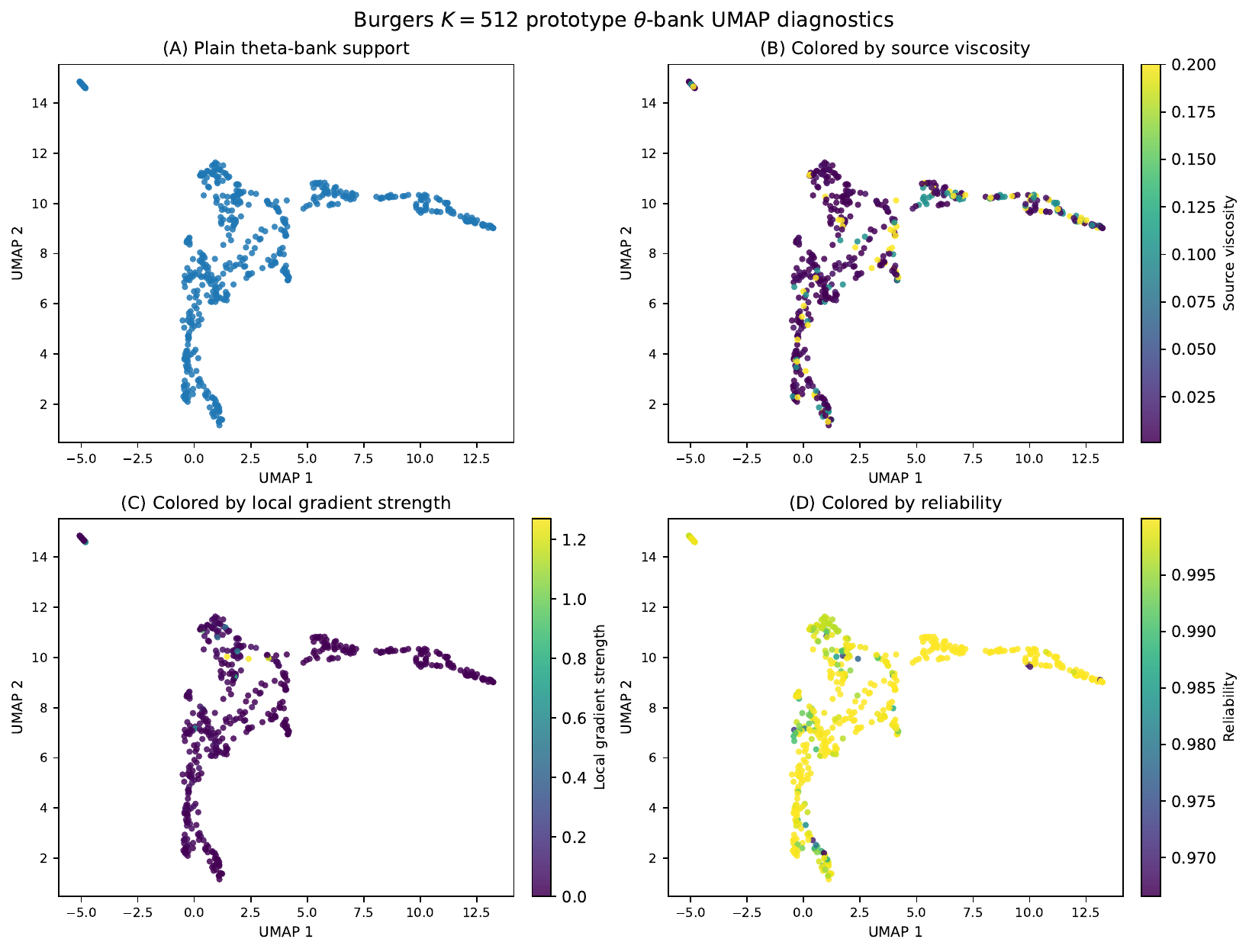}
    \caption{
    \textbf{Burgers prototype \(\theta\)-bank geometry, qualitative visual diagnostics.}
    Two-dimensional UMAP embeddings of the \(K=512\) Burgers prototype \(\theta\)-bank are shown without coloring and with colorings by source viscosity, local gradient strength, and reliability. Each point corresponds to a prototype/support point in the archived \(\theta\)-bank rather than to an individual raw trajectory sample. The panels are intended as qualitative visual diagnostics that complement the quantitative mechanism-space evidence in the main text; they should not be read as proof of intrinsic dimensionality, exact clustering by viscosity, or identification of true physical mechanisms.
    }
    \label{fig:app_burgers_theta_umap}
\end{figure}

Figure~\ref{fig:app_burgers_theta_umap} provides qualitative visual diagnostics for the \(K=512\) Burgers prototype \(\theta\)-bank. These visualizations are included only as a complement to the quantitative mechanism-space diagnostics in the main text.

\FloatBarrier
\clearpage
\section{External Baseline Comparisons and Protocol Notes}
\label{app:external}
\label{app:wb2-protocol-notes}

\subsection{WB2 protocol note}
\label{app:wb2proto}

\paragraph{Protocol.}
Current WeatherBench2 evaluations in this paper use fixed-horizon supervised evaluation, not autoregressive rollout. All \textbf{train=2000} metrics reported for the scarce-data analysis are produced under the same fixed-lead supervised objective and evaluated at fixed +24h and +72h leads on the 6-hour grid.

The \textbf{main external baseline} in the primary WB2 comparison is a 2D Fourier Neural Operator (FNO) trained and evaluated under this protocol. Appendix~\ref{app:wb2-style-baselines} additionally reports lightweight architecture-style stress baselines inspired by AFNO/FourCastNet-style spectral mixing, U-Net/Conv-style residual stacks, and a GraphCast-style message-passing-lite module; these are trained under the same scarce-data fixed-horizon pipeline and should not be read as official pretrained GraphCast, official pretrained Pangu, or global operational NWP systems.

\subsection{Full-data WeatherBench2 horizon boundary scan}
\label{app:wb2-full-data-horizon-boundary}

In the full-data WeatherBench2 horizon scan, the easiest \(+24\)h lead remains a boundary case for the current mechanism-space model, while the Direct--Full gap narrows as the forecast lead increases. This analysis is included as a scope diagnostic rather than as the main real-system evidence. The main text therefore focuses on the stronger scarce-data \(+72\)h setting, while this appendix records the full-data horizon behavior.  The RMSE gap narrows from \(-0.06\) to \(+0.005\) for temperature and from \(-112.5\) to \(-16.2\) for Z500 between \(+24\)h and \(+120\)h. The Full model also exhibits smaller error-growth slopes, corresponding to approximately \(5\%\) and \(9\%\) relative slope reductions for temperature and Z500, respectively.

\begin{figure}[!htbp]
\centering
\includegraphics[width=0.88\linewidth]{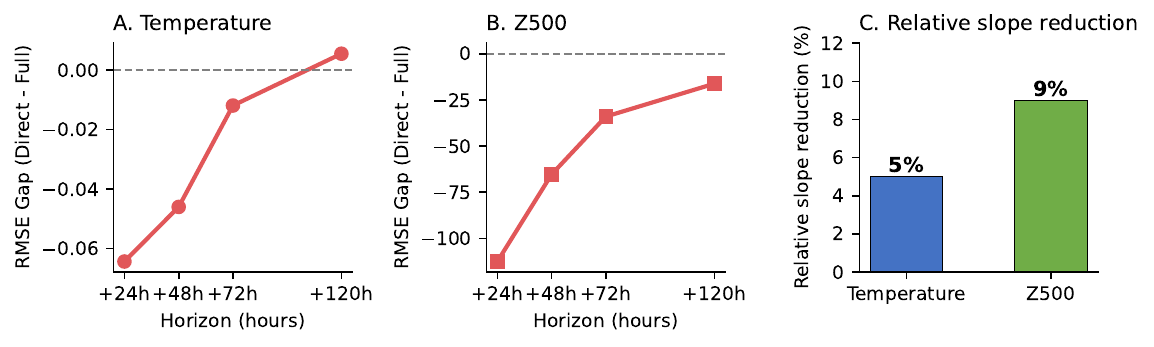}
\caption{
\textbf{Full-data WeatherBench2 horizon boundary scan.} In the full-data setting, the easiest \(+24\)h lead remains a boundary case for the current mechanism-space model, while the Direct--Full gap narrows as the lead increases and the Full model has smaller error-growth slopes. This figure is used as a scope diagnostic; the main real-system evidence in the paper is the scarce-data \(+72\)h comparison.
}
\label{fig:app_wb2_full_data_horizon_boundary}
\end{figure}

\subsection{WB2 external-baseline comparison}
\label{app:wb2ext}
\label{app:wb2-external-baseline}

\textbf{\(+24\)h scope analysis.} Some gains appear at the easier \(+24\)h lead, especially for Z500 versus Direct in paired tests, while other comparisons remain less consistent across variables and baselines. We therefore treat \(+24\)h as a scope analysis and focus the main text on the clearer \(+72\)h scarce-data evidence in Figure~\ref{fig:wb2_72h_main}. The corresponding means and paired tests are reported in Tables~\ref{tab:wb2_summary}--\ref{tab:wb2_paired}.

\textbf{\(+72\)h strongest scarce-data setting.} \(+72\)h is the strongest scarce-data setting in our current WB2 tables: Full shows a clearer, broader advantage over Direct, NoBank, and the external FNO on the temperature and $Z500$ paired summaries (Tables~\ref{tab:b1_wx}--\ref{tab:b2_paired}).
Parameter counts are reported in Table~\ref{tab:b3_wxpt}; in this WB2 comparison, the Full model is substantially smaller than the Direct and External FNO baselines, so the +72h advantage is not attributable to a larger parameter budget.

\begin{table*}[t]
\centering
\scriptsize
\caption{WB2 scarce-data summary, train$=$2000, 64$\times$32, fixed-horizon supervised RMSE (mean $\pm$ std; five seeds). \emph{Current WB2 protocol uses fixed-horizon supervised evaluation, not autoregressive rollout.}}
\label{tab:b1_wx}
\label{tab:wb2_summary}
\begin{tabular}{@{}lcc@{}}
\toprule
\multicolumn{3}{@{}c@{}}{\textbf{+24h}} \\
\midrule
Method & Temp & $Z500$ \\
\midrule
External FNO & $5.9630\pm0.8566$ & $358.6433\pm19.3617$ \\
Direct & $2.0730\pm0.0030$ & $520.2016\pm1.6264$ \\
NoBank & $2.0551\pm0.0560$ & $367.3952\pm20.9635$ \\
Full & $2.0702\pm0.0134$ & $364.8364\pm9.9077$ \\
\addlinespace[0.35em]
\midrule
\addlinespace[0.2em]
\multicolumn{3}{@{}c@{}}{\textbf{+72h}} \\
\midrule
Method & Temp & $Z500$ \\
\midrule
External FNO & $10.5000\pm1.7050$ & $885.3328\pm40.3480$ \\
Direct & $3.2412\pm0.0033$ & $1001.5119\pm7.2978$ \\
NoBank & $3.3498\pm0.1492$ & $985.8954\pm46.0128$ \\
Full & $2.3080\pm0.5155$ & $491.5441\pm221.0308$ \\
\bottomrule
\end{tabular}
\par\smallskip
\footnotesize
\textbf{Note (seed 1024, Full, +72h, temp):} one seed is a reproducible outlier; all five seeds are retained; same-seed rerun matches the originally logged value without replacing the table row.
\end{table*}

\begin{table*}[t]
\centering
\scriptsize
\setlength{\tabcolsep}{2.5pt}
\caption{WB2 paired tests (Full vs.\ baseline), win count and two-sided $p$-value. \emph{Current WB2 protocol uses fixed-horizon supervised evaluation, not autoregressive rollout.}}
\label{tab:b2_paired}
\label{tab:wb2_paired}
\begin{tabular}{@{}l *{8}{c}@{}}
\toprule
 & \multicolumn{2}{c}{\textbf{+24h, T}} & \multicolumn{2}{c}{\textbf{+24h, $Z500$}} & \multicolumn{2}{c}{\textbf{+72h, T}} & \multicolumn{2}{c@{}}{\textbf{+72h, $Z500$}} \\
\cmidrule(lr){2-3} \cmidrule(lr){4-5} \cmidrule(lr){6-7} \cmidrule(lr){8-9}
Comparison & win & $p$ & win & $p$ & win & $p$ & win & $p$ \\
\midrule
Full vs.\ External FNO & 5/5 & $5.3{\times}10^{-4}$ & 2/5 & 0.61 & 5/5 & $5.5{\times}10^{-4}$ & 4/5 & 0.025 \\
Full vs.\ Direct & 2/5 & 0.64 & 5/5 & $2.9{\times}10^{-6}$ & 5/5 & 0.015 & 5/5 & 0.0061 \\
Full vs.\ NoBank & 1/5 & 0.57 & 2/5 & 0.80 & 5/5 & 0.016 & 5/5 & 0.0063 \\
\bottomrule
\end{tabular}
\end{table*}

\begin{table*}[t]
\centering
\scriptsize
\caption{WB2 trainable parameters and training wall time (5 seeds where available). External-FNO wall-clock is reported from the current runs; comparable wall-clock logs are unavailable for the corresponding legacy Direct/NoBank/Full runs. Parameter counts are reported for transparency; the WB2 comparisons are matched in task, data protocol, and evaluation setup rather than strictly parameter-matched.}
\label{tab:b3_wxpt}
\begin{tabular}{@{}lcc@{}}
\toprule
Method & Trainable params & Train time (note) \\
\midrule
External FNO & 2{,}214{,}594 & $214.9\pm149.1$\,s (5 seeds) \\
Direct & 7{,}705{,}922 & N/A \\
NoBank & 400{,}354 & N/A \\
Full & 412{,}738 & N/A \\
\bottomrule
\end{tabular}
\end{table*}

\subsection{Additional WB2 same-protocol architecture-style stress baselines}
\label{app:wb2-style-baselines}

To stress-test whether the scarce-data \(+72\)h result depends only on comparing against Direct, NoBank, and the external FNO, we evaluate several \textbf{lightweight architecture-style} models under the same train=2000, 64\(\times\)32 grid, fixed-horizon WeatherBench2 supervision used for Table~\ref{tab:b1_wx}. These runs share the same evaluation protocol and seed set, but the archived pack trains them with a \textbf{shorter epoch budget} (\texttt{max\_epochs=15}) than the primary Table~\ref{tab:b1_wx} runs; they are therefore best read as \textbf{same-data family stress checks}, not as tight “equal-training-budget” strength references. A Pangu-style patch-transformer variant exists in the same pack but is \textbf{omitted here} because transformer-family baselines are excluded from this submission round.

Table~\ref{tab:wb2_style_baselines} lists means \(\pm\) standard deviations at \(+72\)h. The purpose is \textbf{scoped}: it supports the claim that, within this scarce-data fixed-horizon setting, Full remains strongest among the evaluated models in the table. It does \textbf{not} establish a global WeatherBench2 ranking, nor a comparison to official pretrained GraphCast or Pangu, nor to operational numerical weather prediction.

\input{table_wb2_style_baselines.tex}

\noindent\textit{Provenance:} seed-level and summary CSVs for the style rows are archived with the anonymized code release.

\subsection{Seed 1024 integrity note (Full, +72h)}
\label{app:seed1024}

We document a same-seed Full model +72h integrity rerun under an identical protocol.
The rerun reproduces the original logged run to numerical agreement; \textbf{nevertheless} the original 5-seed rows (including seed 1024) are \textbf{retained unchanged} in the reported tables.
The outlier in 1024 surface temperature is therefore treated as \textbf{reproducible seed-level variance}, not a spurious I/O artifact (see also Table~\ref{tab:b1_wx} footnote).

\clearpage
\subsection{Lorenz96 phase-sweep sweet spots}
\label{app:lorenz96-phase-sweep}

We report the Lorenz96 phase-sweep summaries here because they refine the scope of the method rather than serving as the main evidence in the Results section. Within the scanned forcing values at \(N=16\) and phase-sweep horizon \(h=4\), the relative improvement of Full over Direct is \(11.8\%\) for \(F=6\), \(18.3\%\) for \(F=8\), and \(15.4\%\) for \(F=12\). Within the scanned dimensions at \(F=8\) and the same phase-sweep horizon, the corresponding improvements are \(11.1\%\) for \(N=8\), \(17.8\%\) for \(N=16\), and \(10.0\%\) for \(N=32\). These percentages are computed from the Lorenz96 phase-sweep summaries using \((\mathrm{Direct}-\mathrm{Full})/\mathrm{Direct}\) on the reported \texttt{direct\_test} and \texttt{mech\_test} values. They are not recomputed from the autoregressive rollout RMSE values reported in Table~\ref{tab:b4_lz}.

\begin{figure}[!htbp]
\centering
\includegraphics[width=0.85\linewidth]{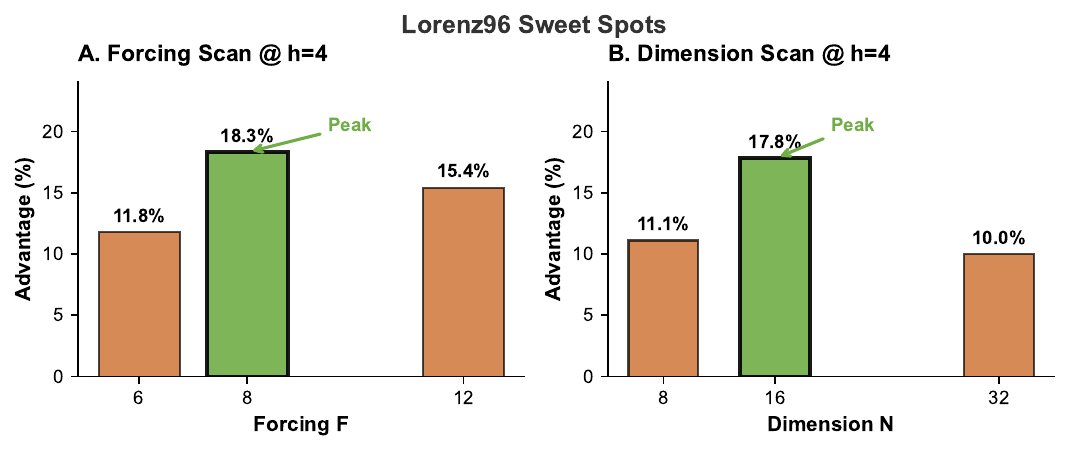}
\caption{
\textbf{Lorenz96 phase-sweep sweet spots.} Relative gains are largest at the tested intermediate forcing and dimension values. This phase-sweep summary is used as a scope analysis for the intermediate-complexity interpretation; it is distinct from the autoregressive rollout and NODE comparisons reported in the appendix.
}
\label{fig:s1}
\label{fig:lorenz_sweetspot}
\end{figure}

\clearpage
\subsection{Lorenz96 continuous-time baseline (NODE)}
\label{app:node}

We include a Neural ODE (NODE) as a continuous-time dynamics baseline, alongside the discrete LSTM. NODE complements, rather than replaces, LSTM. In the sweet-spot configuration ($F{=}8, N{=}16$) and across the autoregressive test horizons summarized in Table~\ref{tab:b4_lz}, the Full model remains strongest relative to Direct, LSTM, and NODE on the reported means; this ordering is specific to the reported sweet-spot configuration and scanned horizons.

\begin{figure*}[!htbp]
\centering
\vspace{2pt}
\includegraphics[width=0.38\textwidth]{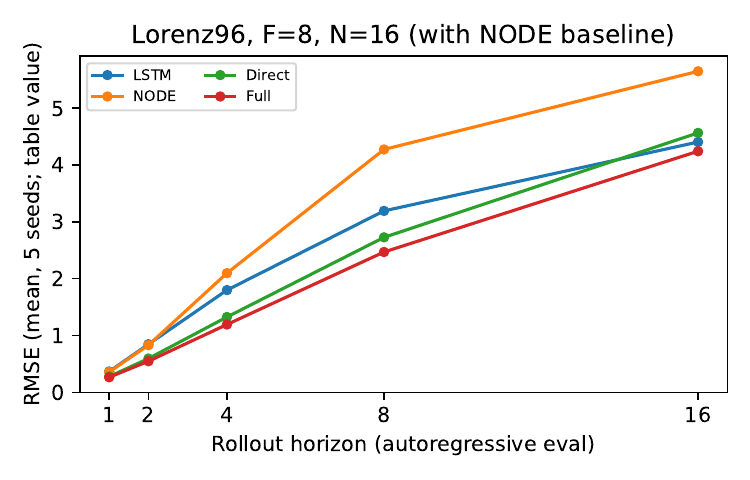}
\hfill
\includegraphics[width=0.56\textwidth]{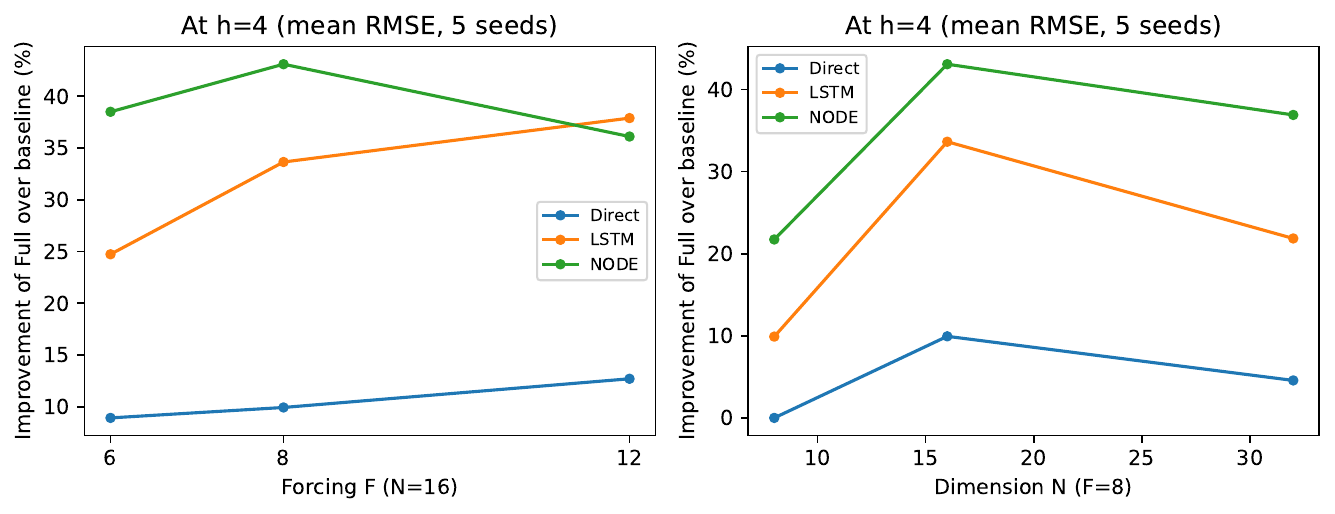}
\caption{\textbf{Lorenz96 with Neural ODE (NODE) baselines.} \textbf{(Left)} Autoregressive RMSE vs.\ rollout horizon at $F{=}8$, $N{=}16$ (mean $\pm$ std over five seeds; Table~\ref{tab:b4_lz}; lines connect means). \textbf{(Right)} Sweet spot at $h{=}4$: percent improvement of Full over each baseline, $(\mathrm{RMSE}_{\mathrm{base}}-\mathrm{RMSE}_{\mathrm{Full}})/\mathrm{RMSE}_{\mathrm{base}}\times 100$ (forcing scan at $N{=}16$; dimension scan at $F{=}8$).
All panels in this figure use the autoregressive rollout evaluation summarized in Table~\ref{tab:b4_lz}, rather than the phase-sweep summary protocol used for Figure~\ref{fig:lorenz_sweetspot}.}
\label{fig:appB_lorenz}
\end{figure*}
\medskip

\begin{table*}[t]
\centering
\scriptsize
\caption{Lorenz96 with NODE: mean autoregressive RMSE (5 seeds) at $h\!\in\!\{1,4,16\}$. Top: forcing scan at $N{=}16$; bottom: dimension scan at $F{=}8$. \textbf{Bold:} lowest RMSE among methods for that ($F$ or $N$) and~$h$.
Here $h$ denotes the autoregressive rollout horizon; this rollout table is a separate evaluation from the phase-sweep percentage summary shown in Figure~\ref{fig:lorenz_sweetspot}.}
\label{tab:b4_lz}
\begin{tabular}{@{}lcccc@{}}
\toprule
\multicolumn{5}{@{}l}{\textbf{Forcing scan} ($N{=}16$)} \\
$F$ & Method & $h{=}1$ & $h{=}4$ & $h{=}16$ \\
\midrule
6 & LSTM & 0.1972 & 0.8766 & 2.5072 \\
6 & NODE & 0.2267 & 1.0727 & 3.7979 \\
6 & Direct & 0.1778 & 0.7245 & 2.6342 \\
6 & Full & \textbf{0.1733} & \textbf{0.6598} & \textbf{2.4619} \\
\midrule
8 & LSTM & 0.3665 & 1.7986 & 4.4042 \\
8 & NODE & 0.3574 & 2.0971 & 5.6482 \\
8 & Direct & 0.2769 & 1.3252 & 4.5646 \\
8 & Full & \textbf{0.2648} & \textbf{1.1935} & \textbf{4.2417} \\
\midrule
12 & LSTM & 0.8003 & 3.8720 & \textbf{6.7914} \\
12 & NODE & 0.6816 & 3.7642 & 7.8394 \\
12 & Direct & 0.5568 & 2.7554 & 7.5641 \\
12 & Full & \textbf{0.5028} & \textbf{2.4052} & 7.0729 \\
\midrule
\multicolumn{5}{@{}l}{\textbf{Dimension scan} ($F{=}8$)} \\
$N$ & Method & $h{=}1$ & $h{=}4$ & $h{=}16$ \\
\midrule
8 & LSTM & 0.1603 & 0.5423 & 2.4463 \\
8 & NODE & 0.1595 & 0.6243 & 3.5766 \\
8 & Direct & \textbf{0.1578} & \textbf{0.4886} & \textbf{2.3603} \\
8 & Full & 0.1601 & \textbf{0.4886} & 2.3643 \\
\midrule
16 & LSTM & 0.3665 & 1.7986 & 4.4042 \\
16 & NODE & 0.3574 & 2.0971 & 5.6482 \\
16 & Direct & 0.2769 & 1.3252 & 4.5646 \\
16 & Full & \textbf{0.2648} & \textbf{1.1935} & \textbf{4.2417} \\
\midrule
32 & LSTM & 0.6062 & 2.6401 & \textbf{4.7534} \\
32 & NODE & 0.6587 & 3.2697 & 5.6296 \\
32 & Direct & 0.4122 & 2.1617 & 4.7087 \\
32 & Full & \textbf{0.3945} & \textbf{2.0630} & 5.2324 \\
\bottomrule
\end{tabular}
\end{table*}

\begin{table*}[t]
\centering
\scriptsize
\caption{Lorenz96: trainable parameters and mean wall-clock training time (seconds). Parameter counts are reported for transparency; the Lorenz96 comparisons are matched in task, data protocol, and evaluation setup rather than strictly parameter-matched.}
\label{tab:b5_lzpt}
\begin{tabular}{@{}lcc@{}}
\toprule
Method & Trainable params & Wall train time (mean $\pm$ std) \\
\midrule
LSTM & 27{,}670 & $7.2\pm0.3$ \\
NODE & 27{,}985 & $23.7\pm0.2$ \\
Direct & 27{,}750 & $6.0\pm0.2$ \\
Full & 59{,}430 & $8.7\pm0.3$ \\
\bottomrule
\end{tabular}
\end{table*}

\clearpage
\subsection{Lorenz96 reservoir stress check}
\label{app:l96-reservoir-stress}

To stress-test the Lorenz96 sweet-spot comparison against strong non-neural and reservoir-family baselines, we add RC-ESN and NVAR under the same \(F{=}8, N{=}16\) autoregressive rollout protocol used in Table~\ref{tab:b4_lz}. Reservoir computing has been widely used for model-free prediction of spatiotemporally chaotic systems, making it a natural stress baseline for this setting~\citep{pathak2018model}. Before evaluating these additional baselines, we verified that the means for Direct, LSTM, NODE, and Full in the sweet-spot \((F{=}8,N{=}16)\) block of Table~\ref{tab:b4_lz} are reproduced by the same evaluator and data generator to within \(5\times 10^{-4}\). The stress baselines are therefore reported as an exact-protocol supplement to the reported Lorenz96 sweet-spot comparison, not as a separate benchmark setting.

\begin{table}[t]
\centering
\small
\caption{Lorenz96 reservoir stress check at the intermediate-complexity sweet spot \((F{=}8,N{=}16)\). Values are mean autoregressive RMSE at horizons \(h\in\{1,4,16\}\). Lower is better. RC-ESN and NVAR are added as exact-protocol stress baselines; Direct, LSTM, NODE, and Full are the Table~\ref{tab:b4_lz} sweet-spot means (five seeds).}
\label{tab:l96_reservoir_stress}
\begin{tabular}{@{}lccc@{}}
\toprule
Method & \(h{=}1\) & \(h{=}4\) & \(h{=}16\) \\
\midrule
LSTM & 0.3665 & 1.7986 & 4.4042 \\
NODE & 0.3574 & 2.0971 & 5.6482 \\
Direct & 0.2769 & 1.3252 & 4.5646 \\
RC-ESN & 0.2979 & 1.3734 & 4.3671 \\
NVAR & 2.347 & 4.289 & 46.64 \\
Full & \textbf{0.2648} & \textbf{1.1935} & \textbf{4.2417} \\
\bottomrule
\end{tabular}
\end{table}

Full has the lowest RMSE across the reported horizons in this stress-check table. Relative to RC-ESN, Full reduces autoregressive RMSE by approximately \(11.1\%\), \(13.1\%\), and \(2.9\%\) at \(h=1,4,16\), respectively. This supports the scoped statement that the Full model achieves state-of-the-art-level forecasting performance in the tested intermediate-complexity Lorenz96 setting. We do not claim global Lorenz96 state of the art: the statement is restricted to the reported \(F{=}8,N{=}16\) autoregressive rollout setting and the baselines in Tables~\ref{tab:b4_lz} and~\ref{tab:l96_reservoir_stress}.

\section{Related Work}
\label{app:extended-related-work}

This appendix expands the positioning summarized in the Introduction. Existing scientific forecasting methods can be organized around several related responses to the same difficulty: future states are often easier to predict when additional structure beyond the raw state is available. Standard data-driven weather forecasting treats prediction primarily as state-to-state forecasting; physics-aware and continuous-time models impose or exploit known dynamical structure; operator-learning and latent-dynamics methods learn structured evolution maps or hidden dynamical processes; and prototype, retrieval, or memory-based methods provide reusable representational support. These directions all introduce structure beyond raw state regression. The limitation addressed in this paper is different: they usually do not formulate forecasting as online inference of the currently active local mechanism in an empirical mechanism space. Mechanism-space forecasting makes that object explicit by mapping observed histories to local mechanism descriptors, organizing those descriptors into an empirical predictive geometry, and forecasting through the inferred active mechanism within that geometry.

\paragraph{Terminology in this appendix.} This appendix follows the terminology of the main text: the central object is the empirical mechanism space, while prototypes are treated as support points or anchors within that space. When comparing to prototype, memory, or retrieval methods, the relevant distinction is therefore not whether a bank is used, but whether forecasting is formulated as inference over an empirical geometry of local evolution rules.

\subsection{Direct state forecasting in scientific systems}

Modern scientific forecasting is dominated by direct state prediction, in which a model maps the current observed state, or a short history of states, directly to a future state. This paradigm underlies much of the recent progress in data-driven weather forecasting, including FourCastNet, Pangu-Weather, GraphCast, and NeuralGCM, which differ substantially in architecture but all operate primarily as state-to-state forecasters. WeatherBench~2 further standardized this setting by providing a common benchmark for comparing AI and numerical weather forecasting systems across medium-range horizons \citep{Pathak2022FourCastNet,Bi2023PanguWeather,Lam2023GraphCast,Kochkov2024NeuralGCM,Rasp2024WeatherBench2}.

These models have made direct state prediction the dominant evaluation framing in modern scientific forecasting, especially in medium-range weather benchmarks. The dominant evaluation framing in this literature remains model-centric: the question is typically which architecture predicts future states more accurately. Our work addresses a different question: when should the forecasting coordinate itself shift from observed states to inferred local mechanisms?

\subsection{Latent dynamics, operator learning, and mechanism representations}

Our work is also related to a broader line of research that seeks intermediate dynamical structure rather than purely static input-output mappings. Neural ODEs and Latent ODEs introduced continuous-time hidden dynamics as a way to represent temporal evolution through an underlying latent dynamical process. In parallel, operator-learning methods such as DeepONet and Fourier Neural Operator shifted attention from pointwise regression to learning evolution operators over functions and fields. These lines of work demonstrate that forecasting and dynamical modeling can benefit from representations that are more structured than a conventional feedforward state encoder \citep{Chen2018NeuralODE,Rubanova2019LatentODE,Lu2021DeepONet,Li2021FNO}.

Data-driven dynamical-systems methods also provide useful context. Dynamic mode decomposition extracts coherent linear evolution structure from observations, Koopman-based predictors lift nonlinear dynamics into coordinates with approximately linear evolution, and sparse identification methods seek parsimonious governing equations from data~\citep{schmid2010dynamic,korda2018linear,brunton2016discovering}. These approaches aim to recover or exploit dynamical structure, whereas our mechanism space is used operationally as an empirical predictive geometry for online mechanism inference.

Meta-learning methods such as Model-Agnostic Meta-Learning (MAML) also seek reusable structure across tasks, but they do so by learning parameter initializations for rapid adaptation rather than by constructing an explicit mechanism space; in particular, they do not formulate forecasting as inference over an empirical mechanism space with anchored finite support \citep{Finn2017MAML}.

Our mechanism-learning framework is related in spirit to these approaches, but differs in emphasis. We do not merely introduce another latent state. Instead, we explicitly construct and query a mechanism space built from localized dynamical descriptors, and use this space as the intermediate object through which forecasting is performed. In this sense, our contribution is not only a new forecasting architecture, but also an empirical study of when forecasting through a structured mechanism representation is preferable to forecasting directly in state space.

\subsection{Prototype, retrieval, and memory-based inductive biases}

A second line of related work concerns the use of structured support sets, prototypes, or external memories to organize learned representations. Matching Networks and Prototypical Networks showed that representing classes through support examples or prototype vectors can provide a powerful and simple inductive bias, especially in low-data settings. These methods operate in classification rather than scientific forecasting, but they establish an important precedent: retrieval and prototype structure can improve generalization by constraining inference to organized regions of representation space rather than allowing fully unconstrained predictions \citep{Vinyals2016MatchingNetworks,Snell2017ProtoNets}.

Our use of prototype support differs from these methods in both object and role. The prototypes in our framework do not represent semantic classes; they represent recurring regions of a learned mechanism space. Retrieval therefore does not serve label transfer, but mechanism anchoring: it constrains the inferred mechanism to remain near previously observed dynamical templates. This distinction is central to our results, especially because our NoBank ablation tests whether anchored mechanism inference matters beyond a generic latent pathway: the gain comes not merely from adding an intermediate mechanism pathway, but from finite support in mechanism space that anchors that pathway to recurring dynamical templates.

\subsection{Benchmarks, predictability, and mechanism-space forecasting}

Our experimental design spans three complementary settings. WeatherBench~2 provides a realistic benchmark for medium-range weather forecasting. Burgers dynamics sits naturally within the broader scientific-machine-learning benchmark tradition represented by PDEBench, while Lorenz96 serves as a classic controlled chaotic system for studying predictability and regime-dependent behavior. Together, these systems allow us to probe not only raw forecasting performance but also structure, stability, and variation across dynamical regimes \citep{Rasp2024WeatherBench2,Takamoto2022PDEBench,Lorenz1996Predictability}.

Our results connect to recent analyses of long-horizon instability in neural operators by showing that anchored mechanism-space forecasting provides a complementary perspective to direct state forecasting \citep{McCabe2024StabilityOperators}.

%% file: table_wb2_style_baselines.tex
% WB2 architecture-style stress baselines table (generated from archived scarce-data pack CSVs + tab:b1_wx + provenance note for authors).
\begin{table*}[t]
\centering
\scriptsize
\setlength{\tabcolsep}{3pt}
\caption{Additional WeatherBench2 \textbf{same-protocol, train=2000, 64$\times$32, fixed-horizon} architecture-style \textbf{stress} baselines at \textbf{+72h} (RMSE mean $\pm$ std; five seeds). \emph{Not} official pretrained GraphCast/Pangu, \emph{not} operational NWP, and \emph{not} a global WeatherBench2 leaderboard claim. Primary four rows match Table~\ref{tab:b1_wx}. Style rows come from the archived scarce-data pack with \textbf{shorter training} (\texttt{max\_epochs=15}) than the primary runs (see appendix text). \textbf{Omitted:} Pangu-style patch-transformer (transformer family excluded this round).}
\label{tab:wb2_style_baselines}
\begin{tabular}{@{}l >{\raggedright\arraybackslash}p{0.19\textwidth} >{\raggedright\arraybackslash}p{0.24\textwidth} cc c >{\raggedright\arraybackslash}p{0.13\textwidth}@{}}
\toprule
\textbf{Method} & \textbf{Type} & \textbf{Protocol (abbr.)} & \textbf{Temp} & \textbf{$Z500$} & \textbf{\#} & \textbf{Claim use} \\
\midrule
Full & Proposed & train=2000; fixed-horizon sup.; +72h & $2.3080\pm0.5155$ & $491.5441\pm221.0308$ & 5 & Main \\
Direct & Matched internal baseline & same & $3.2412\pm0.0033$ & $1001.5119\pm7.2978$ & 5 & Main \\
NoBank & Ablation & same & $3.3498\pm0.1492$ & $985.8954\pm46.0128$ & 5 & Main \\
External FNO & External FNO baseline & same & $10.5000\pm1.7050$ & $885.3328\pm40.3480$ & 5 & Main \\
\midrule
AFNO-style (spectral-lite) & Arch.-family stress baseline & same data/eval/seeds; shorter train budget & $31.5621\pm21.0367$ & $1020.2936\pm2.8061$ & 5 & Appendix only \\
U-Net pack & Arch.-family stress baseline & same & $3.2467\pm0.0061$ & $1020.5826\pm0.3382$ & 5 & Appendix only \\
GraphCast-style low-data & Arch.-family stress baseline & same & $53.8088\pm35.5858$ & $1022.9266\pm4.9634$ & 5 & Appendix only; not official GC \\
\bottomrule
\end{tabular}
\end{table*}